\documentclass[letterpaper, 10 pt,  conference]{IEEEtran}
\usepackage{xcolor,soul,framed} 
\usepackage{color, soul}
\IEEEoverridecommandlockouts                             
\usepackage[letterpaper,top=57pt,bottom=43pt,left=48pt,right=48pt]{geometry}

\usepackage{amsmath}
\usepackage{amssymb}
\usepackage{amsbsy}
\usepackage{comment}
\usepackage{bm}
\usepackage{pifont}
\usepackage{algorithm2e}
\RestyleAlgo{ruled}
\DeclareRobustCommand\blackcircle[2]{\tikz[baseline=(char.base)]{\node[shape=circle,draw,fill=black, text=white,inner sep=0pt] (char) {\textit{#1}\textsubscript{\fontsize{4}{5}\selectfont #2}}}}
\DeclareRobustCommand\blackcircleSt[2]{\tikz[baseline=(char.base)]{\node[shape=rectangle,minimum width=0.3cm, minimum height=0.3cm, draw,fill=black, text=white,inner sep=0pt] (char) {#1}}}
\DeclareRobustCommand\circleSt[1]{\tikz[baseline=(char.base)]{\node[shape=rectangle,minimum width=0.3cm, minimum height=0.3cm, draw,inner sep=0pt] (char) {#1}}}
 
\SetCommentSty{mycommfont}
\SetKwComment{Comment}{$\triangleright$\ }{}
\usepackage{orcidlink}
\usepackage{tikz}
\usepackage{tcolorbox}
\newcommand{\remove}[1]{\ignorespaces}

\begin{document}
    \title{Attention-based Estimation and Prediction of Human Intent to augment Haptic Glove aided Control of Robotic Hand}
  \author{Muneeb~Ahmed\orcidlink{0000-0003-0323-8260}, Rajesh~Kumar, Qaim Abbas, Brejesh Lall, Arzad A. Kherani,  Sudipto Mukherjee
  \thanks{Manuscript received July 25, 2023. Manuscript revised November 1, 2023. \textit{(Corresponding author: Muneeb Ahmed.)}}
  \thanks{Muneeb Ahmed, Qaim Abbas, Brejesh Lall, Arzad Kherani,  and Sudipto Mukherjee are with Indian Institute of Technology Delhi, Hauz Khas, New Delhi-110016, India (e-mail: muneeb.ahmed@dbst.iitd.ac.in, qaim.abbas@ee.iitd.ac.in, brejesh@ee.iitd.ac.in, sudipto@mech.iitd.ac.in).}
  \thanks{Rajesh Kumar is with Addverb Technologies Bot Valley, Noida-201310, India  (e-mail: rajesh.kumar01@addverb.com).}
  }

\markboth{IEEE ROBOTICS AND AUTOMATION LETTERS, VOL.XX, NO.XX, JULY~2023
}{Muneeb \MakeLowercase{\textit{et al.}}: Estimation and Attention-based Prediction of Human Intent to augment Haptic Glove aided Control of Robotic Hand}
\maketitle
\begin{abstract}
The letter focuses on Haptic Glove (HG) based control of a Robotic Hand (RH) executing in-hand manipulation of certain objects of interest. The high dimensional motion signals in HG and RH possess intrinsic variability of kinematics resulting in difficulty to establish a direct mapping of the motion signals from HG onto the RH. An estimation mechanism is proposed to quantify the motion signal acquired from the human controller in relation to the intended goal pose of the object being held by the robotic hand. A control algorithm is presented to transform the synthesized intent at the RH and allow relocation of the object to the expected goal pose. The lag in synthesis of the intent in the presence of communication delay leads to a requirement of predicting the estimated intent. We leverage an attention-based convolutional neural network encoder to predict the trajectory of intent for a certain lookahead to compensate for the delays. The proposed methodology is evaluated across objects of different shapes, mass, and materials. We present a comparative performance of the estimation and prediction mechanisms on 5G-driven real-world robotic setup against benchmark methodologies. The test-MSE in prediction of human intent is reported to yield $\sim$97.3$\mathbf{-}$98.7\% improvement of accuracy in comparison to LSTM-based benchmark. 
\end{abstract}
\IEEEpeerreviewmaketitle
\section{Introduction}
\IEEEPARstart{T}{\lowercase{eleoperation}} systems have been a focus of research for decades\cite{colasanto2012hybridWHYNOTDIRECTMAPPING}. In recent years, teleoperation in robots have developed across various domains, including tele-surgery \cite{cheng2021neural}, space exploration \cite{ wang2020event}, industry \cite{owan2020fastermanufacturing}. The aim is to control a remotely placed robot \cite{qin2022onegraspprecision} based on the intention of the human. Numerous studies have explored analysis, estimation and prediction of the signals transmitted during teleoperation for pick and place modules with serial robotic arms, in-hand manipulation maneuvers constituting precision grasp modes are limited \cite{handa2020dexpilot}. Dexterous in-hand manipulation \cite{andrychowicz2020learning, cruciani2020benchmarking} recruits motion of the fingertips of a robotic hand (RH) such that the object moves relative to the palm. Although in-hand manipulation has visual similarity to the traditional pick and place operations using serial robots, the former demands precise control of the motion of each of the fingertips \cite{qin2022onegraspprecision} and may ensure constraints on the force applied by the fingertips in order to ensure that the object does not slip from the grip. Difficulty is encountered in mapping the combination of the two measures, force and motion, from the human hand to the RH. This is compounded by the RHs developed having kinematics different from that of the human hand \cite{colasanto2012hybridWHYNOTDIRECTMAPPING}. 
\par For the case of the RHs, multiple fingertips combine to manipulate the object. The motion of the fingers of the human hand is captured at intervals via a haptic glove (HG). The motion commands to the joints of the RH lead to a finite motion of the fingertips. Although the HG motion signal represents the motion of the joints of the human fingers, the signal usually does not represent the complete object kinematics. As the kinematic architecture of the RH is seldom similar to that of the human hand, there is no one-to-one mapping between the two signals. Apart from the kinematic variance of the two devices, the interaction of the two with real-world objects is also dissimilar. The two manipulator-object interactions differ in contact stiffness and local friction characteristics, necessitating different interpretations of the signals in discussion. Although the signal from the HG contains essential information about the joint motion, the same cannot be transmitted directly to the RH. Instead, an appropriate encoding-decoding and subsequent control strategy are needed to transform the signal from the human into an interpretable human intent. Then, a subsequent strategy should be deployed to interpret the human intent signal at the RH and issue an appropriate control signal in order to complete the desired map. \begin{figure*}
    \centering
    \includegraphics[width =0.8\linewidth]{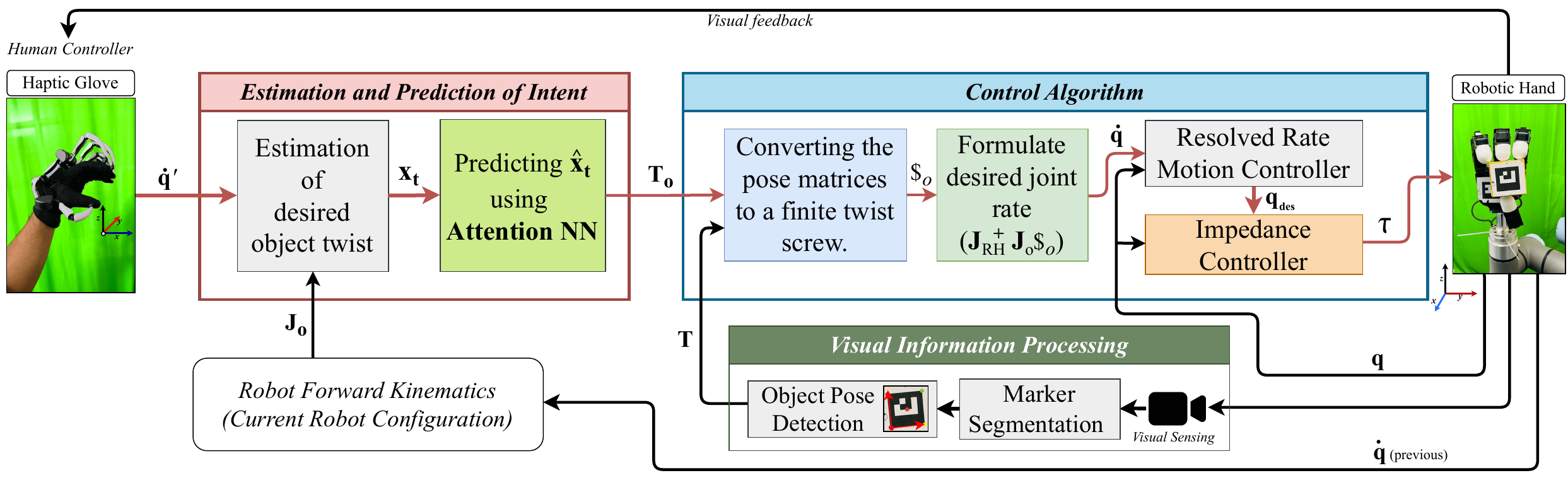}
    \caption{The proposed system workflow for estimation \& the prediction of human intent with corresponding control mechanism. The estimation mechanism (left) is detailed in Sections \ref{sec:estimationintent} and \ref{sec:predictedintent}; The control algorithm (right) is detailed in Section \ref{sec:controlsection}.}
    \label{ctral}
    \vspace{-0.5cm}
\end{figure*}
It shall also be detailed later that due to a significant initial acceleration phase in the human motion, the motion of the robot tends to lag the motion of the human user. This effect is multiplied when the teleoperation is performed over a distance. Hence, a learning algorithm is proposed to generate a human intent signal based on the measurements to mitigate the overall latency. \par In this work, design of workflow breaks down into a two-fold substructure viz., \textit{estimation and prediction of intent}, and \textit{control algorithm}. The primal reasons in using this design strategy are as a consequence of analysing the signals emerging from haptic glove, and also ensuring that the effect of communication delay during teleoperation is mitigated. We observed explained variance ratios from the principal component decomposition of HG signals to have significance up to the first six principal components, which could be taken as indicators of the pose of the object being manipulated, given that six parameters (degrees of freedom) are necessary and sufficient to represent the state of a rigid body in Cartesian space. Hence, the estimation mechanism ensures the transformation of complex HG signal into a signal that defines the estimated pose of the object (lesser in dimension). Further, the prediction mechanism utilizes neural network approach to synthesize a prediction value of the estimated pose. A predicted value will help mitigate the effect of latency observed during teleoperation. Finally, the control mechanism transforms the estimated/predicted pose onto the joint space of the robot. The granular details of these algorithms are stated in Section \ref{sec:methodology}.
In summary, the contributions of the work are stated as follows:
\begin{itemize}
    \item Estimation of the desired goal pose of the object being manipulated by the RH based on the HG joint motion.
    \item Synthesis of a control algorithm such that RH achieves the joint configuration as defined by the estimated goal pose of the object.
    \item Leveraging attention-based convolutional neural network (CNN) encoder for predicting of the goal pose based on an intent template to mitigate latency in remote teleoperation scenario.
\end{itemize}

\section{Recent Works}
    Teleoperation systems for in-hand manipulation essentially comprise a human controller (local site) and a robotic hand (remote site)  \cite{li2022transteleopMARKERLESS}. The measurement of movements relayed from the human controller can be observed by a plethora of sensing mechanisms such as vision \cite{qin2022onegraspprecision, handa2020dexpilot}, kinaesthetic data \cite{lii2012towardGESTURE}, exoskeleton devices \cite{liu2017gloveEXPOSKELETONHAPTICGLOVE}, and hybrid approaches \cite{colasanto2012hybridWHYNOTDIRECTMAPPING, low2017hybridGLOVEEXOSKELETONHAPTICGLOVE}.  The approach of 3D hand pose estimation and transformation of its pose for robot demonstration is widely recognized. [DIME] demonstrated a marker-based approach for identification of key points in the human hand to estimate its pose, leverage an imitation learning algorithm to train dexterous manipulation policies, followed by subsequent transformation into the joint configuration space of the robotic hand. Some studies leverage direct-mapping \cite{peer2008multiDIRECTMAPPING} strategy for achieving successful retargeting. However, such algorithms are not generalizable due to their significant coupling with the robotic hardware and the morphological mismatch in the kinematics between the human controller and the robot \cite{colasanto2012hybridWHYNOTDIRECTMAPPING}. In another approach, Dex-Pilot \cite{handa2020dexpilot} presented a strategy to perform in-hand manipulation of various objects, focusing on estimating a template of human hand and capturing its joint motion in a constrained space using visual cues. This eliminates the use of marker-based pose estimation but introduces a heavily parameterized pose estimation mechanism that may incur significant processing delays. Secondly, achieving precision in pose estimation from vision-based input is coupled with environmental lighting conditions, noise, and occlusions.  In specific environments where modelling visual input tends to be volatile, kinematic and exoskeleton-based hybrid approaches \cite{colasanto2012hybridWHYNOTDIRECTMAPPING} can yield better results. Synthesis of policies on visual cues necessitate complex modelling. Hence, their respective inference times are significantly high. Further, the controller's environment is limited by the field of view of the camera. Due to this complex arrangement, it is not always desired to establish a teleoperation setup utilizing visual cues as human input for the policies especially in scenarios where latency is critical. It is noteworthy to mention that retargeting algorithms deployed at the remote site also depend on the communication delay in the channel that relays the estimated pose or raw input from the user (depending on the placement of the pose estimation algorithm). In this context, images/videos that correspond to high-dimensional data utilize a significant amount of bandwidth. Hence, it is desired to utilize low-dimensional human input for a certain estimation algorithm towards in-hand manipulation such as kinaesthetic data captured with the help of a haptic glove. Keeping in consideration the complexity in teleoperation of multi-fingered robots, ALOHA \cite{zhao2023learningALOHA} presented a  hardware system for bimanual teleoperation specifically improvising the precision in manipulation tasks using imitation learning with action chunking mechanism over human demonstrations. We infer from their work that utilizing action chunking with transformers on low-dimensional input reduces the training time and increases the test performance of the precision control. They reported an inference time of 0.01 sec. None of the work reported in this survey addresses to the channel latency. Our work addresses both these challenges of latency and camera dependence. Furthermore, our work addresses the retargeting of human pose onto the robot joint configuration via an intent template, which is a task-relevant term that helps bridge the high-level objectives of the controller and the low-level control of the robot, bringing about an abstraction. It is later seen that prediction on the intent template yields the best performance against the benchmarks in the literature.  Since agile in-hand rotation/spinning remains a primal task for demonstration of teleoperation and learning for a plethora of recent works such as Transteleop \cite{li2022transteleopMARKERLESS}, Dextreme \cite{handa2023dextreme}, and others \cite{ANDREWmorgan2022complex,9744489YiLiuVR, 10004028osherazulay,yang2023tacgnnLEARNINGTACTILEGNN}. We compare our proposed approach against these works for in-hand rotation task. In summary, our proposed approach for in-hand manipulation aided by a haptic glove is motivated by addressal to the kinematic difference between the controller and the robot, latency issue in the channel, and precise prediction and control.

\section{Methodology}
\label{sec:methodology}
The proposed system architecture consists of two subsystems, the RH and the HG (as shown in Fig. \ref{ctral}). The human user wears HG, which senses the human motion at $a_k$ joints such that $a_k \leq a_h$, where $a_h$ is the number of finger joints on the human hand which exactly characterises the motion of all the fingers. It means that the HG does not precisely characterise the motion of the human fingertips, like in the case of Dexmo Haptic Glove (DHG) \cite{gu2016dexmo} that senses the motion signal at 11 points on the human hand as opposed to 19 odd joint actuation in the human fingers.  Also, the HG restricts motion of the human fingers at $a_g$ joints, such that $a_g \leq a_k$ (using a restrictive torque). The other subsystem is the RH consisting of $a_r$ active joints with $p$ serial chains out of the tree type robotic system \cite{shah2013dynamics}. The RH and the HG have different kinematic morphology and might be differently oriented (for instance, during the experiment, the Allegro RH is right oriented whereas the DHG is left-oriented). The only similarity between the two robotic systems seems to be that the two are tree type robotic systems with serial chains. We aim however to map the intent of the human using the HG and deploy control signals to the RH to achieve the intended manipulation. The details of intent estimation, prediction, and retargeting of intent template onto the robot joint configuration are specified in Sections \ref{sec:estimationintent},  \ref{sec:predictedintent}, and \ref{sec:controlsection}, respectively.
\subsection{Feedback-based Control Architecture for Sticking Manipulation of the Object}
\label{sec:controlsection}
\label{control}
\begin{figure}[ht]
    \centering
    \includegraphics[width=0.6\linewidth,trim={6cm 5.5cm 9cm 3.5cm},clip]{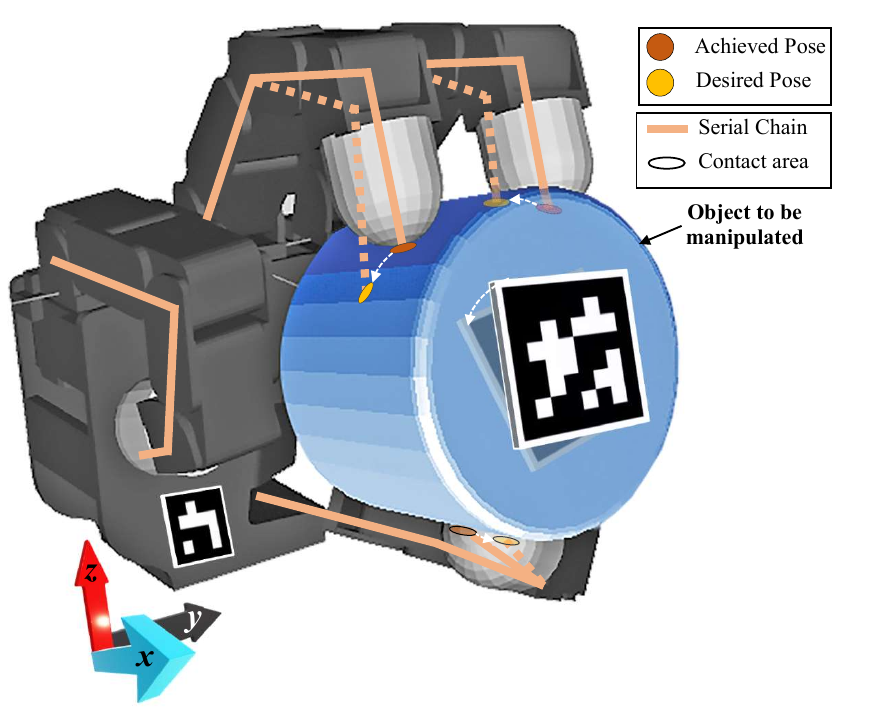}
    \caption{Illustration of a tree-type robotic hand holding an object of interest to undergo in-hand manipulation from the current/achieved pose to the desired pose.}
    \label{impedance}
    \vspace{-0.55cm}
\end{figure}

Consider an RH with $a_r$ actuated joints and $p$ serial chains within the tree (Fig. \ref{impedance}). Each of the serial chain needs to interact with the object and perform manipulation such that the goal (target) object pose is reached. The manipulation considered in this letter is a rotation of the object. 
Additionally, the $a_r$ actuated joints are torque/ current-controlled. Let $\mathbf{r}_i$ be the vector connecting a suitable point on the object to the $i^{th}$ ($1 \leq i \leq p)$ contact point on the surface of the object. It is known that the equations governing forces that can be applied by the fingertips of the RH form an under-specified set of  equations, with null space solutions\cite{kumar1988force}.
\begin{equation}
   \underbrace{\left[\begin{array}{cccc}\mathbf{1}_{3\times3}&\mathbf{1}_{3\times3}&...&\mathbf{1}_{3\times3} \\
    \mathbf{\tilde{r}_1}&\mathbf{\tilde{r}_2}&...&\mathbf{\tilde{r}_p}\end{array}\right]}_{\mathbf{G}}\underbrace{\left[\begin{array}{c}\mathbf{f_1}\\\mathbf{f_2}\\\vdots\\\mathbf{f_p}\end{array}\right]}_{\mathbf{F}} = \bm{\mathcal{W}}
    \label{map}
\end{equation}
where $\mathbf{\tilde{r}}_i$ refers to the cross product matrix corresponding to the vector $\mathbf{r}_i $, and $\mathbf{1}_{3\times3}$ represents the $3\times3$ identity matrix. The vector $\mathbf{f}_i$ refers to the force applied by the fingertips on the surface of the object by the $i^{th}$ fingertip. The vector $\bm{\mathcal{W}} \in \mathbb{R}^6$ refers to the object dynamics (net forces and moments applied to the object) by the fingertips \cite{davidson2004robots} represented as  $\bm{\mathcal{W}} = \left[\begin{array}{c}\mathbf{M}_o\mathbf{a}_o \\ \mathbf{I}_o\bm{\dot{\omega}}_o + \bm{\omega}_o\times\mathbf{I}_o\bm{\omega}_o\end{array}\right]
$. The term $\mathbf{M}_o$ is the mass of the object, $\mathbf{a}_o$ the linear acceleration of the object, $\mathbf{I}_o$ the moment of inertia tensor of the object, and $\bm{\omega}_o$ is the angular velocity of the object. It is known that the solution to the grasp map (Eq. \ref{map}) is a combination of two vector fields. The fingertip forces belonging to the equilibrating force field imparts motion to the object. However, the interaction force field~\cite{kumar1988force} generates the null solutions. Though motion is not imparted, assigning suitable null forces leads to improvement of the contact condition.\\\indent It is also known that the force vectors belonging to the interaction force field can be spanned using the vectors joining the fingertips. Consider the case shown in Fig. \ref{impedance}, which is a three-point grasp. The general solution to the force  field \cite{zhang1996determination, kumarenhanced} is usually represented as $\mathbf{F} = \mathbf{G}^+\bm{\mathcal{W}} + \sum_{i=1}^{3p-6}\alpha_iN(\mathbf{G})$, where $p>2$, $N(\mathbf{G})$ represents the null space of the matrix $\mathbf{G}$ (the grasp matrix as defined in Eq. \ref{map}), $\mathbf{G}^+$ represents the Moore-Penrose Pseudo Inverse \cite{barata2012moore} of the matrix $\mathbf{G}$, and $\alpha_i$ represent the coefficients of the vectors belonging to $N(\mathbf{G})$. The dynamics of the RH, as with all kinematic chains, with the vector $\mathbf{q}$ as the joint state of the robot is generally represented as $\mathbf{M}\mathbf{\ddot{q}} + \mathbf{C(\mathbf{q,\dot{q}})} + \mathbf{g(q)} = \bm{\tau} - \mathbf{J}^T\mathbf{F}_{\text{ext}}$, where $\mathbf{M}$ refers to the mass matrix of the robot, $\mathbf{C}$ refers to the Coriolis term of the dynamics, $\mathbf{g(q)}$ refers to the gravity compensation term of the robot, $\bm{\tau}$ vector refers to the torque applied at the actuated joints. The vector $\mathbf{F}_{\text{ext}}$ is the force applied to the end effectors. Derivation for dynamics of tree-type robots in the same form is available as well \cite{shah2013dynamics}.  
Consider the robot grasping the object with individual fingertips controlled under impedance control. The torque vector $\bm{\tau}$ is determined as
  $  \bm{\tau} =\mathbf{K}(\mathbf{q}_{\text{des}} - \mathbf{q}) + \mathbf{D}(\mathbf{\dot{q}}_{\text{des}} - \mathbf{\mathbf{\dot{q}}}) + \mathbf{M}\mathbf{\ddot{q}}_{\text{des}} + \mathbf{C}(\mathbf{q}_{\text{des}}, \mathbf{\dot{q}}_{\text{des}}) + \mathbf{g}(\mathbf{q}_{\text{des}})$
where $\mathbf{q_{\text{des}}}$ is the desired motion of the joints required to achieve the desired position. The matrices $\mathbf{K}$, $\mathbf{D}$ are the gain matrices for robotic control \cite{siciliano2016springer}. Since, any general rigid body motion is represented as an element of the $\mathbf{SE}(3)$ manifold. Therefore, a pose matrix defining the state of the object can be represented as 
 $\mathbf{T} \equiv $~$ \{\left[\begin{array}{cc}\mathbf{R}&\mathbf{o}\\ \mathbf{0}&1\end{array}\right] \mid \mathbf{R}\in \mathbb{R}^{3\times3}, \mathbf{o} \in \mathbb{R}^3, \mathbf{R}^T\mathbf{R} = \mathbf{1}_{3\times3}, $~$ det(\mathbf{R}) = 1\}$. The twist screw $\mathbf{\bm{\$}}_{o} \in \mathbb{R}^6$ represents the twist representation (combination of the linear and the angular velocity) relating the current object pose ($\mathbf{T}$) to the desired object pose ($\mathbf{T_o}$).  The trajectory of screw rotation, at the object level, denoted by $\bm{\$}_o$, defines the transition of the object from its current position to the intended goal position. The twist of the end-effectors of the RH, is represented as $\mathbf{\bm{\$}_{RH}} = \mathbf{J_{RH}}\mathbf{\dot{q}}$, where $\mathbf{J_{RH}}$ denotes the Jacobian Matrices that establish the relationship of joint rates ($\mathbf{\dot{q}}$) of the active joints in RH to the observable twist in the end-effectors of RH. Also, $\mathbf{\bm{\$}_{RH}} = \mathbf{J_{o}}\mathbf{\bm{\$}_{o}}$, where $\mathbf{\bm{\$}_{o}}$ denotes the twist in the object, and Jacobian matrices ($\mathbf{J_{o}}$) relate the twist in the object to the observable twist ($\mathbf{\bm{\$}_{RH}}$) in the RH's end-effectors.  Utilising the standard resolved rate motion approximation, the subsequent incremental goal pose is determined using the equations,$\mathbf{\dot{q}_{des}} = \mathbf{J_{RH}^+}\mathbf{\bm{\$}_{RH}}$ and $\mathbf{q} = \mathbf{q} + \eta\mathbf{\dot{q}}\Delta t$
where, $\eta \in \mathbb{R}$ controls the incremented value to the current joint configuration towards the desired goal pose in the time interval ($\Delta t$). In this whole process, the object twist is derived from the desired object pose matrix ($\mathbf{T_o}$) and the current object pose ($\mathbf{T}$) matrix. A separate vision-based subsystem calculates the matrix $\mathbf{T}$ utilizing a position sensitive marker such as ArUco. Upon calibrating this subsystem offline, the object's true/current pose is determining by segmenting the marker on the object and calculating its relative angle with respect to the marker on the RH, in real-time. Having established a control algorithm to ensure that the robotic fingertips result in the RH to achieve the desired object pose, we focus on the synthesis of the goal object pose from the signals from the human wearing the HG. A pseudo-code representation of the control algorithm is described in Algorithm \ref{algo:controlalgo}.
\begin{algorithm}[hbt!]
\SetKwInput{KwData}{Input}
\SetKwInput{KwResult}{Output}
\DontPrintSemicolon
\label{algo:controlalgo}
\caption{Synthesis of Control Mechanism for RH}
\KwData{Current Object Pose (i.e., Ground Truth) from camera in the form of Transformation Matrix $\mathbf{T}$; Jacobian of object (i.e., $\mathbf{J_o}$); Goal Transformation Matrix (i.e., $\mathbf{T_o}$)}
\KwResult{Joint Rate of Robotic Hand (i.e., $\mathbf{\dot{q}}$)}
\While{True}{
$\mathbf{J}_{\text{RH}} \gets getJacobian(\text{RH})$ \Comment*[r]{Jacobian of Robotic Hand}
$\mathbf{R} \gets \mathbf{T}_{[1:3,1:3]}$  \Comment*[r]{Sliced Rotation Matrix (Ground Truth)}
$\mathbf{R_o} \gets \mathbf{T_o}_{[1:3,1:3]}$  \Comment*[r]{Sliced Rotation Matrix (Predicted)}
$\bm{\omega}_{\text{desired}} \gets log(inverse(\mathbf{R})\mathbf{R_o})$ \Comment*[r]{Matrix logarithm}
$\mathbf{\bm{\$}}_{\text{RH}} \gets \mathbf{J_o}\mathbf{\bm{\$}_o}$  \Comment*[r]{ $\bm{\$}_{\text{RH}}$ is the twist of end effectors of RH}
$\mathbf{\dot{q}} \gets pseudoinverse(\mathbf{J}_{\text{RH}})\mathbf{\bm{\$}}_{\text{RH}}$  \Comment*[r]{$\mathbf{\dot{q}}$ is joint rate of RH}
}
\end{algorithm}
\subsection{Estimation of the Human Intent by Superposition of Serial Chain Kinematics}
\label{sec:estimationintent}
As noted earlier, the only kinematic similarity between the RH and the HG is the tree-type robotic system with serial chains. Further, the joint encoders on the HG do not precisely characterise the exact motion of the human fingertips. A heuristic is proposed for the superposition of the kinematics of the serial chains of the HG to the kinematics of the serial chains of the RH. The overall architecture and data flow to relate the input signal on the HG to the motion of the object on the RH is presented schematically in Fig. \ref{ctral}.  Consider the mapping of the fingertip of the $i^{th}$ serial chain of RH and the $i^{th}$ serial chain of the HG. The first-order differential kinematics of the $i^{th}$ fingertip of the RH is represented as $\mathbf{\bm{\$}}_i = \sum_{j = 1}^{a_r}\$_{ij}\dot{q}_j$, where $\$_{ij}$ represents the screw rotation of the $j^{th}$ joint of the $i^{th}$ chain. The kinematics of the $i^{th}$ serial chain, including the active joints of the HG on the $i^{th}$ serial chain yields the  hypothetical first order kinematics of the hand as $\mathbf{\bm{\$}}_i' = \sum_{j = 1}^{k}\$'_{ij}\dot{q}'_j$. The vector $\mathbf{\bm{\$}}_i'$ represents the equivalent screw motion of the active joint on the HG to the RH. For all the $k$ hypothetical kinematics of the HG on the RH, the estimated object screw motion is characterised as the minimum norm rigid body motion, quantified as $\mathbf{\bm{\$}'_o} = \mathbf{J_o}^+\mathbf{\bm{\$}}_{\text{HG}}$. The vector $\mathbf{\bm{\$}}_{\text{HG}}$ is quantified as, $\mathbf{\bm{\$}}_{\text{HG}} =  \left[\begin{array}{ccc}\mathbf{\bm{\$}}'_1 &  \hdots & \mathbf{\bm{\$}}'_p \end{array}\right]^T$. The instantaneous linear velocity and the angular velocity of a point on the object can be derived based on the screw representation as $[\mathbf{v},~\mathbf{\bm{\omega}}] = \mathbf{\bm{\$}'_o}$. 
Upto a time $t$, knowing the hypothetical twist motion, the motion of the object can be synthesised as $\mathbf{d} = \int_{0}^t\mathbf{v}dt$, and $\bm{\theta} = \int_{0}^t\mathbf{\bm{\omega}}dt$. Subsequently, the goal transformation matrix is repeatedly synthesised as $\mathbf{T_o} = \left[\begin{array}{cc}\mathbf{R}&\mathbf{d}\\ \mathbf{0}&1\end{array}\right]$, with the matrix $\mathbf{R} = \mathbf{1}_{3\times3}+\mathbf{E}\sin{\theta}+\mathbf{E^2}(1-\cos{\theta})$. The matrix $\mathbf{E}$ is defined as $\mathbf{E} = \left[\begin{array}{ccc}0&-\theta_z&\theta_y\\\theta_z&0&-\theta_x\\-\theta_y&\theta_x&0\end{array}\right]$, where $\bm{\theta} = \left[\begin{array}{c}\theta_x\\ \theta_y\\   \theta_z\end{array}\right]$, $\theta = ||\bm{\theta}||_2$. In this formulation, the integral effect of the hypothetical twist screw is used to repeatedly determine the human intention to manipulate the object. Often the intent of prehensile manipulation is primarily a rotation, in which case the velocity terms $\mathbf{v}$ are equated to $\mathbf{0}$. The pseudo-code representation of this formulation is detailed in Algorithm \ref{algo:synthesisofintent}.
\begin{algorithm}[hbt!]
\SetKwInput{KwData}{Input}
\SetKwInput{KwResult}{Output}
\DontPrintSemicolon
\label{algo:synthesisofintent}
\caption{Estimation and Prediction of Human Intent}
\KwData{Joint Rate of HG (i.e., $\mathbf{\dot{q}'}$); Jacobian of object (i.e., $\mathbf{J_o}$)}
\KwResult{Desired Goal Pose for RH in terms of Goal Transformation Matrix (i.e., $\mathbf{T_o}$)}
$\bm{d} \gets \mathbb{R}^{3 \times 1}$ \Comment*[r]{Displacement vector}
$\bm{\theta} \gets \mathbb{R}^{3 \times 1}$ \Comment*[r]{Angular displacement vector}
$\mathbf{x}_{t} \gets \mathbb{R}^{r\times 1}$ \Comment*[r]{Input vector to prediction mechanism}
$\mathbf{\hat{x}}_{t} \gets \mathbb{R}^{m\times 1}$ \Comment*[r]{Output predicted vector}
$dt \gets c$ \Comment*[r]{$dt$ is a constant, implying a very short interval of time}
\While{True}{
$\mathbf{J}_{\text{HG}} \gets getJacobian(\text{HG})$ \Comment*[r]{Jacobian of Haptic Glove}
$\mathbf{\bm{\$}}_{\text{HG}} \gets \mathbf{J}_{\text{HG}}\mathbf{\dot{q}'}$ \Comment*[r]{Calculating the twist of HG}
$\mathbf{\bm{\$}'_{o}} \gets \text{pseudoinverse}(\mathbf{J_o})\mathbf{\bm{\$}}_{\text{HG}}$ \Comment*[r]{Estimating object twist}
$\mathbf{v} \gets \mathbf{\bm{\$}'_{o}}_{[1:3]}$ \Comment*[r]{3 velocity components in object twist}
$\bm{\omega} \gets \mathbf{\bm{\$}'_{o}}_{[4:6]}$ \Comment*[r]{3 angular velocity components in object twist}
$ \mathbf{d} \gets \mathbf{d} + \mathbf{v}dt $ \Comment*[r]{Integration of displacement}
$ \bm{\theta} \gets  \bm{\theta} + \mathbf{\bm{\omega}}dt$ \Comment*[r]{Integration of angular displacement}
$\mathbf{x}_{t} = \left\{\theta_{x}(t-r+1), \theta_{x}(t-r), \hdots,  \theta_{x}(t)\right\}$ \Comment*[r]{$\mathbf{x}_{t}$ holds previous $r$ values of $\theta_{x}$ upto time $t$ using sliding window}
$\mathbf{\hat{x}}_{t} \gets f(\mathbf{x}_{t},m)$ \Comment*[r]{$\mathbf{\hat{x}}_{t}$ is the predicted vector from the attention model $f()$}
$\theta_{pr} \gets \mathbf{\hat{x}}_{t}[t+m]$ \Comment*[r]{Predicted future intent}
Generate $\mathbf{T_o} $ \Comment*[l]{Section III}
}
\end{algorithm}
\begin{figure*}[ht]
    \centering
    \includegraphics[width = 0.9\linewidth ]{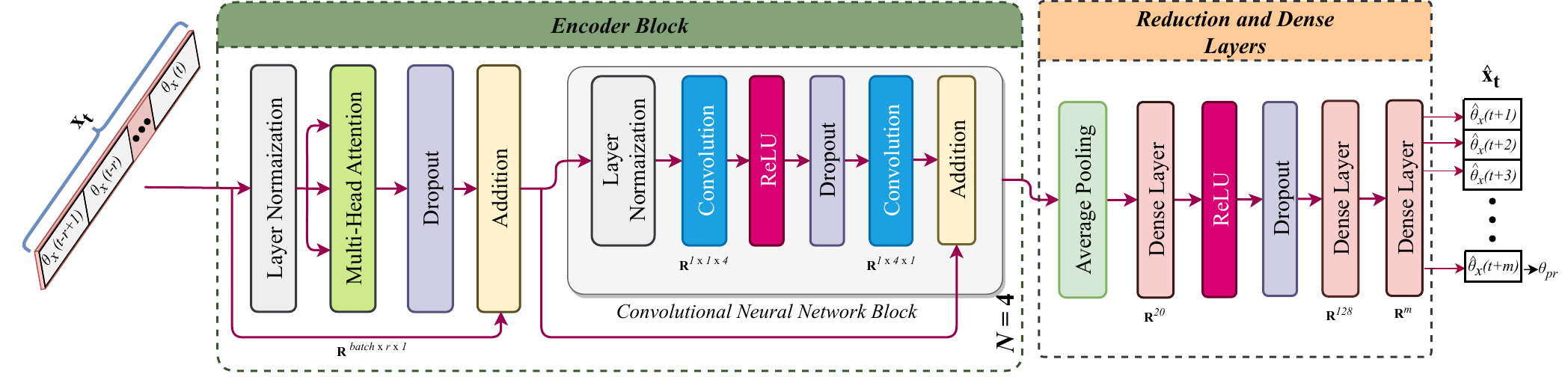}
    \caption{Attention-based proposed architecture for Intent prediction.}
    \label{fig:MODELTRANSFORMER}
    \vspace{-4mm}
\end{figure*}
It shall be seen that the human intent coded as per this methodology accumulates the effect of the instantaneous motion of the human finger joints. We introduce the observed knowledge that the human tendency is to follow a sigmoidal behavior while manipulating objects in hand. This includes an acceleration phase associated with only a small cumulative rotation (or displacement of the hypothetical/real object), followed by a large displacement period and then a small motion and large deceleration period. At the same time, the RH joints require a finite time to accelerate to follow the goal trajectory. The cumulative effect of the human intent, the goal pose, is evident only after the deceleration phase is initiated. This leads to a lag in the motion of the RH vis-\'a-vis, the signal captured by HG. The time delay between the feedback at the HG and visual feedback of the RH is counter-intuitive to a teleoperator.
\subsection{Prediction of the Intended Goal Pose}
\label{sec:predictedintent}
An end-to-end teleoperation of robots deployed in critical applications requires a tolerable communication delay in the order of 250$-$1000 ms \cite{atecs2019design,zhou2023modified1000ms}. To minimise the effect of control latency and any other marginal delays,  an attention-based prediction strategy is introduced in the estimation algorithm, that is effective to preserve the temporal features in the human behavior. We propose to use the heuristic that the humans deploy prehensile manipulation in repeated tasks, giving rise to the necessity of sequence learning.  The architectural specifications of the proposed attention-based CNN encoder to augment the estimation mechanism of human intent of motion. The encoded motion of the joints of the human fingers (formulated as intent of the human operator) is the motion of the object being manipulated. Here, the trajectory of same motion signal of the object is formulated as a sequence.  Consider the estimated angular displacement of the object ($\bm{\theta}$) along a certain axis at the current time instance ($t$) as $\theta_{x}(t)$.  For a motion of the object with respect to a coordinate frame, the time-distributed series $\mathbf{x}_{t} \in \mathbb{R}^r$ consists of previous $r \in \mathbb{Z}^+$ elements sampled sequentially from the signal (using sliding window approach), denoted as $\mathbf{x}_{t} = \left\{\theta_{x}(t-r+1), \theta_{x}(t-r), \hdots, \theta_{x}(t)\right\}$ for $t \geq r$.  Thus, the encoder examines a sequence of preceding $r$ samples from the signal in order to generate an approximation  ${\mathbf{\hat{x}}_{t}} \in \mathbb{R}^m$  of the desired outcome (where $r$  corresponds to the input window size and $m\in \mathbb{Z}^+$ corresponds to the size of the output lookahead vector $\mathbf{\hat{x}}_{t}$). This can be considered for all axes of rotation (depending upon the need). The value $r=20$ is empirically chosen and the length of the output vector/lookahead (i.e., $m$) is varied in each experiment (during training) to obtain an optimal point where the predictions yield the least test-error. Here, we do not explicitly add any positional embedding due to the inherent sequential nature of the input. However, convolutional layers are added after the attention block to capture spatial variance. The convolutional filters tend to extract the variance in across the sub-samples in its input. Since, the output from the encoder block is a time-distributed 1-dimensional (1D) sequence for a particular Cartesian axis, the spatial variance captured correlates to temporal variance. Hence, convolutional operations on the given sequences enrich the temporal features captured by the attention-based backbone. Finally, the encoder block is cascaded $N$ times (the value of $N$ is empirically chosen during experimentation). The output is processed by two cascaded fully connected layers to yield $(t+m)^\text{th}$ predicted value. 
\section{Results}
\subsection{Experimental Framework and Setup}
\begin{figure}[ht]
    \centering
    \includegraphics[width =0.8\linewidth, trim = {0cm 7.2cm 0cm 0.5cm}, clip]{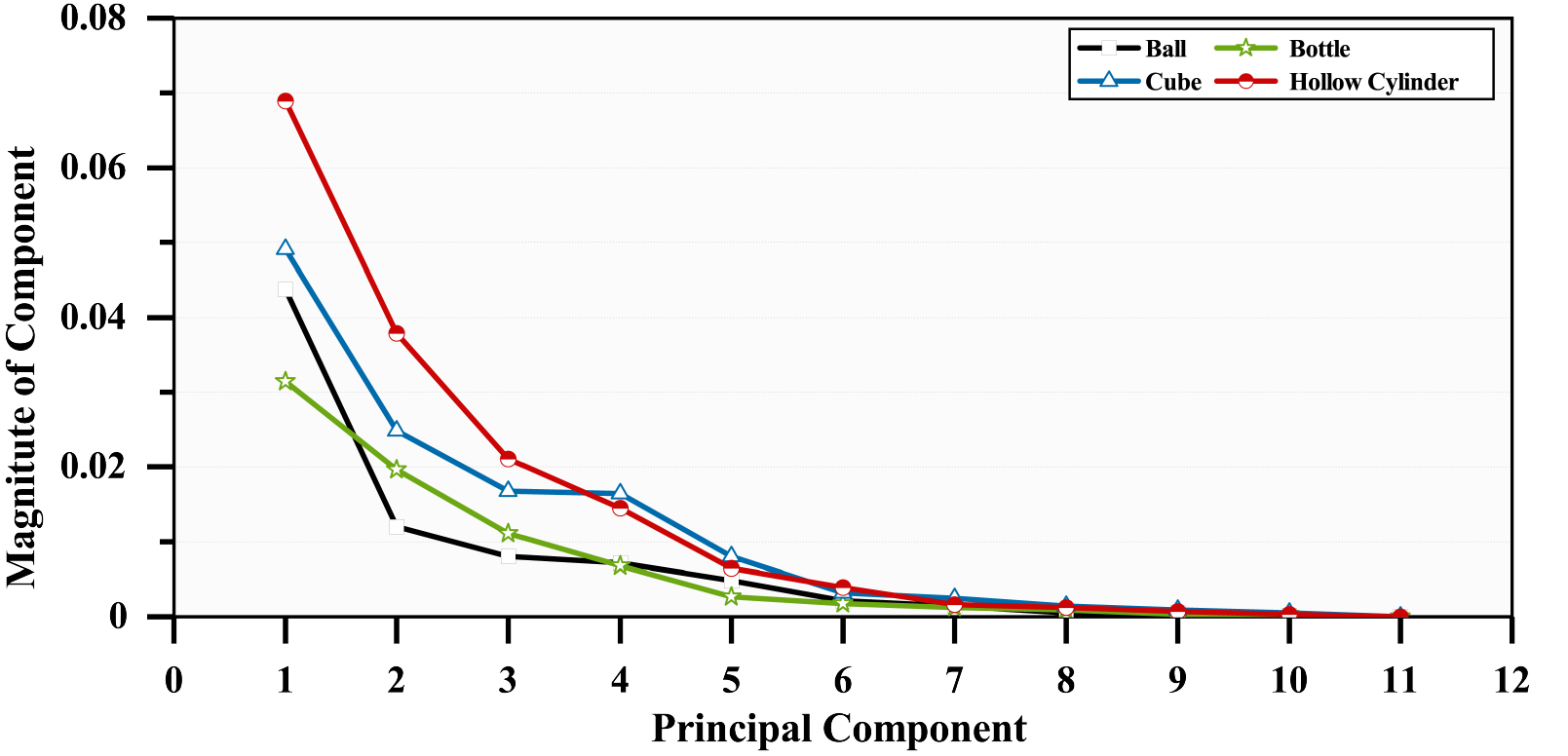}
    \caption{PCA analysis of the data corresponding for exemplar manipulation of different objects.}
    \label{pca_net}
\end{figure}

The demonstration of the results utilizing the proposed mechanisms for intent estimation, prediction and synthesis of control is performed on Allegro Robotic Hand (ARH) that is remotely controlled by a human controller wearing Dexmo Haptic Glove (DHG), as shown in Fig. \ref{ctral}. ARH consists of four kinematic chains (three fingers and a thumb). Each of the kinematic chain has 4 active joints powered by motors with gear ratio of 1:369. At this ratio, the end effector dynamics can be neglected. The impedance controller produces a desired torque at 333 Hz, whereas the resolved rate motion controller updates the desired joint angle value at 20 Hz. In this work, the end effector stiffness is kept constant throughout the manipulation though it could vary with the task. On the other hand, DHG is a tree-type chain of five serial chains (four fingers and a thumb). Each of the fingers has two encoding points, whereas the thumb has three encoding points. The fingers have one encoder recording the bending action while the other encoders record the splitting motion. The variation of the joint values for a test motion of the DHG manipulating certain objects registers a time-distributed 11D signal. The datasamples from these signals along with the estimated representation of the intent (in terms of the angle observed by the object being manipulated) are curated to form a dataset. The proposed design produces a 6-D transformed signal as a result of the estimation and prediction mechanisms, that is transformed back to the joint space of the ARH by the control mechanism. The design choice of 6-dimensions is taken from the reference of observing the following results on application of Principal Component Analysis (PCA): In order to analyse the joint signals from the DHG, PCA decomposition was performed on the input signal from DHG by manipulating different real-world objects. Fig. \ref{pca_net} illustrates the magnitude of PCA decomposition of the DHG signal. 
It can be seen that for all the cases, the first six components are dominant. This could be attributed to the fact that the primary motion targetted of the object during in-hand manipulation was a general rotational motion with parasitic translation motion. Beyond six dimensions, the magnitude of the principal components is negligible, which is consistent with the motion being of a rigid body (3 angular velocity, and 3 linear velocity). Hence, 6-D signal would capture the dynamics of the object as quantification of the user's intent.
\subsection{Results from Intent estimation and Control algorithms}
Being high-dimensional representations, there is no direct mapping of the signals from the HG onto the RH. The embodiment of a non-linear mapping mechanism established, as a result of such intent estimation mechanism, yields a approximation of the desired goal pose of the object estimated from the motion signals procured by DHG. 
\begin{figure*}[ht]
    \centering
    \includegraphics[width = 1\linewidth]{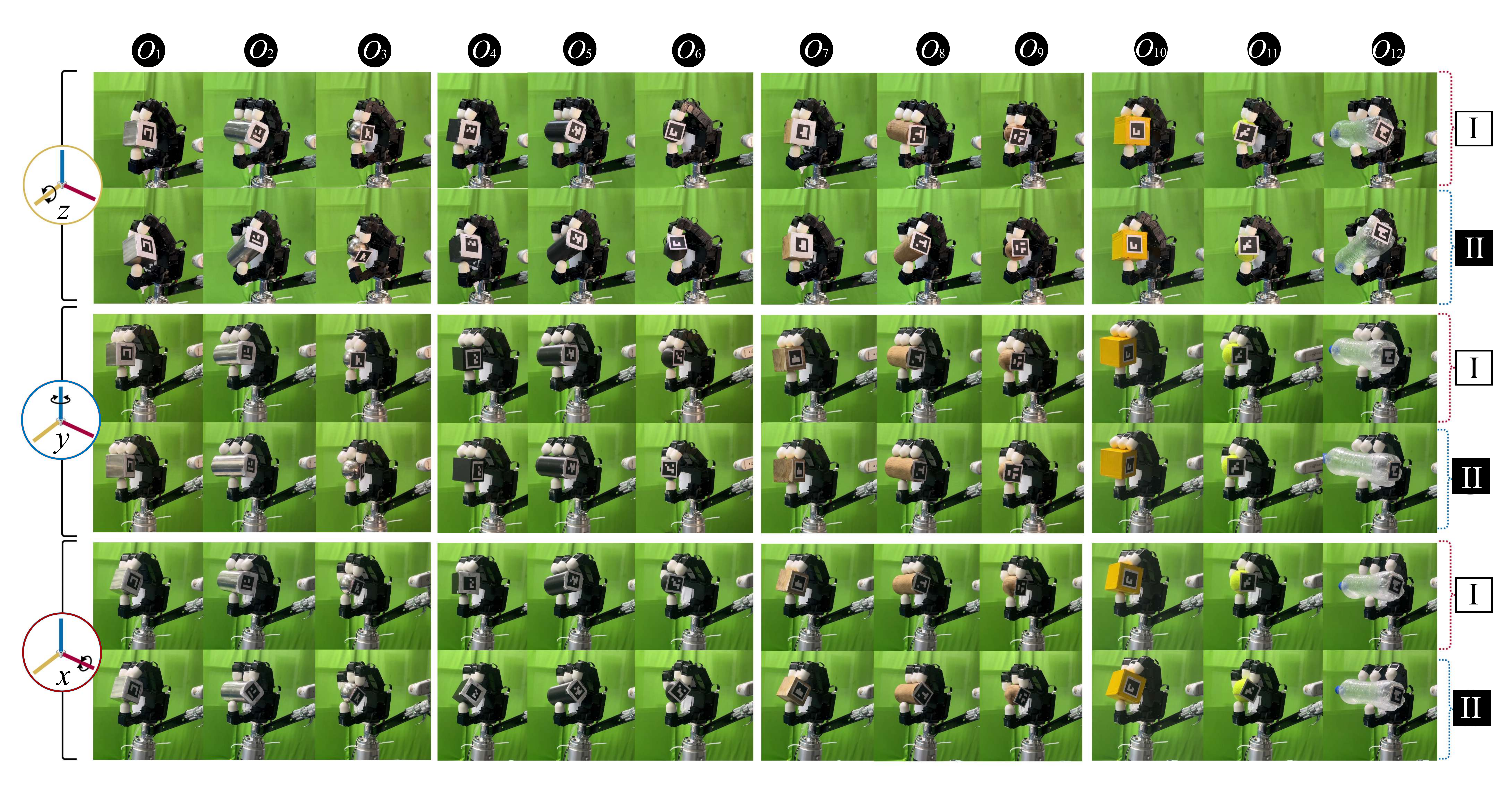}
    \caption{Motion achieved on ARH as a result of estimation, and control mechanisms while manipulating various objects under study about the three Cartesian axes. First two rows show snapshots of pose achieved during rotation about $x$-axis, the next two rows show pose achieved while rotating about $y$-axis and the last two rows show pose achieved while rotating about $z$-axis achieved as a result of proposed methodology while moving from initial pose (\circleSt{I}) towards desired goal pose through intermediate pose (\blackcircleSt{II}~). The results show rotations about the three Cartesian axes using multiple types of objects viz., \blackcircle{O}{1} (Hollow metal cube), \blackcircle{O}{2} (Hollow metal cylinder), \blackcircle{O}{3} (Hollow metal sphere), \blackcircle{O}{4} (Solid plastic cube), \blackcircle{O}{5} (Hollow plastic cylinder), \blackcircle{O}{6} (Solid plastic sphere), \blackcircle{O}{7} (Solid wooden cube), \blackcircle{O}{8} (Solid wooden cylinder), \blackcircle{O}{9} (Solid wooden sphere), \blackcircle{O}{10} (Hollow cardboard cube), \blackcircle{O}{11} (Rubber ball), and \blackcircle{O}{12} (Cylindrical transparent plastic bottle), illustrating the robustness of the proposed approach across various shapes and materials of objects. 
}
    \label{fig:objectmotionresults}
    \vspace{-0.1cm}
\end{figure*}
The subsequent control algorithm transforms the estimated intent signal of human operator into appropriate joint state configuration on the ARH to replicate the motion. The visual results from the proposed estimation and control mechanism in capturing the rotational intent (in one of the three Cartesian directions) during manipulation of three different objects are shown in Fig. \ref{fig:objectmotionresults}. As seen from these results, the overall objective of estimating human intent of rotation while performing object manipulation task is achieved. Further, the extent of rotation of the held object reaches till the grip is maintained and the centre of gravity of the object does not change drastically. It is even possible that the object might fall from the grasp of the robot if the shift in the center of the gravity by the motion disturbs the equilibrium. Further studies are needed to address such challenges.
\subsection{Results from Attention-based prediction mechanism}
It is experimentally observed that the motion of the object by the ARH lags the motion of the HG owing to the delay in processing, control, and communication. The proposed attention-based network trained is then introduced, to predict the goal pose proactively and recursively for mitigation of delays. The model is trained on the data in batches of 16 using Adam \cite{kingma2014adam} optimization with mean-squared error cost function for 200 epochs with a learning rate of 0.0001. Fig. \ref{fig:TRANSFORMERRESULTS} illustrates the outcome of the prediction mechanism for determining the trajectory of the object's pose against the ground truth. A comparative analysis of the proposed methodology against existing studies is illustrated in Table \ref{tab:comparison}. 
\begin{table*}[]
\centering
\caption{COMPARATIVE ANALYSIS THE PROPOSED PREDICTION APPROACH}
\label{tab:comparison}
\begin{tabular}{llllll}
\hline
\textbf{Ref.}                                  & \textbf{Interface (Methodology)}               & \begin{tabular}[c]{@{}l@{}}\textbf{Estimation/}\\ \textbf{Prediction Error}\end{tabular} & \begin{tabular}[c]{@{}l@{}}\textbf{Estimation/}\\ \textbf{Prediction Time}\end{tabular} & \textbf{Teleoperation} & \begin{tabular}[c]{@{}l@{}}\textbf{Latency}\\ \textbf{Addressal}\end{tabular} \\ \hline
TransTeleop\cite{li2022transteleopMARKERLESS}  & Vision (Generative Model)                      & 0.03 rad                                                                                 & 0.027 sec                                                                               & \checkmark                     & \ding{55}                                                                            \\
Dextreme\cite{handa2023dextreme}               & Vision (Sim2Real, Imitation)                   & $\sim$ 0.09 rad                                                                          & 0.06 sec                                                                                & \ding{55}                    & \ding{55}                                                                            \\
Dexpilot\cite{handa2020dexpilot}               & Vision (Neural Network)                        & 0.02 rad per joint                                                                       & 0.03 sec                                                                                & \checkmark                     & \ding{55}                                                                           \\
Osher et. al\cite{10004028osherazulay}         & Haptic (LSTM)                                  & 0.06 - 0.37 rad                                                                          & -                                                                                       & \ding{55}                     & \ding{55}                                                                            \\
ALOHA\cite{zhao2023learningALOHA}              & Vision (Transformer)                           & -                                                                                        & 0.01 sec                                                                                & \ding{55}                    & \ding{55}                                                                           \\
AnyTeleop \cite{qin2023anyteleop}              & Vision (Neural Network)                        & 12.50\%                                                                                  & 0.014 sec                                                                               & \checkmark                    & \ding{55}                                                                            \\
Andrew et. al\cite{ANDREWmorgan2022complex}    & Finger gaiting (Pose estimation)               & 0.118 rad                                                                                & 11.275 sec                                                                              & \ding{55}                   & \ding{55}                                                                            \\
Yi Liu et. al \cite{9744489YiLiuVR}            & VR + Tactile (Direct mapping)                  & 0.114 rad                                                                                & -                                                                                       & \checkmark                     & \ding{55}                                                                           \\
TacGNN \cite{yang2023tacgnnLEARNINGTACTILEGNN} & Tactile (Graph NN)                             & 0.16 rad                                                                                 & -                                                                                       & \ding{55}                     & \ding{55}                                                                            \\
\cite{codebase}                                & Kinaesthetic (LSTM)                            & 0.018 rad                                                                                & 0.003 sec                                                                               & \checkmark                     & \checkmark                                                                           \\
\textbf{\textbf{Ours (Proposed)}}              & \textbf{\textbf{Kinaesthetic (Attention CNN)}} & \textbf{\textbf{0.00047 rad}}                                                            & \textbf{\textbf{0.002 sec}}                                                             & \textbf{\checkmark} & \textbf{\checkmark}                                                        \\ \hline
\end{tabular}
\end{table*}
\section{Discussion}
\indent\textbf{Ablation Study:} An ablation study was performed to vary the length of the output vector  
 ($\mathbf{\hat{x}}_t \in \mathbb{R}^m$, $1 \leq m \leq 20$) to check the performance on the prediction mechanism. Hence, the nodes in the last dense layer ($m$) are varied. The pose matrix $\mathbf{T_o}$ is framed using the value corresponding to the ($t+m$)$^{\text{th}}$ element of the predicted output vector. The effect of changing the dimensions of the output is observed empirically. Variation of the predicted value $\bm{\hat{\theta}}(t+m)$ from the ground truth value $\bm{\theta}(t+m)$ about a certain axis $\alpha$ at time $t+m$ is quantified as MSE, denoted by $\mathcal{L}_\alpha(\hat{\theta}_\alpha,\theta_\alpha)$, formulated as \par
\begin{figure}[ht]
    \centering
    \includegraphics[width = 1\linewidth, trim = {0cm 0cm 1cm 0cm}, clip]{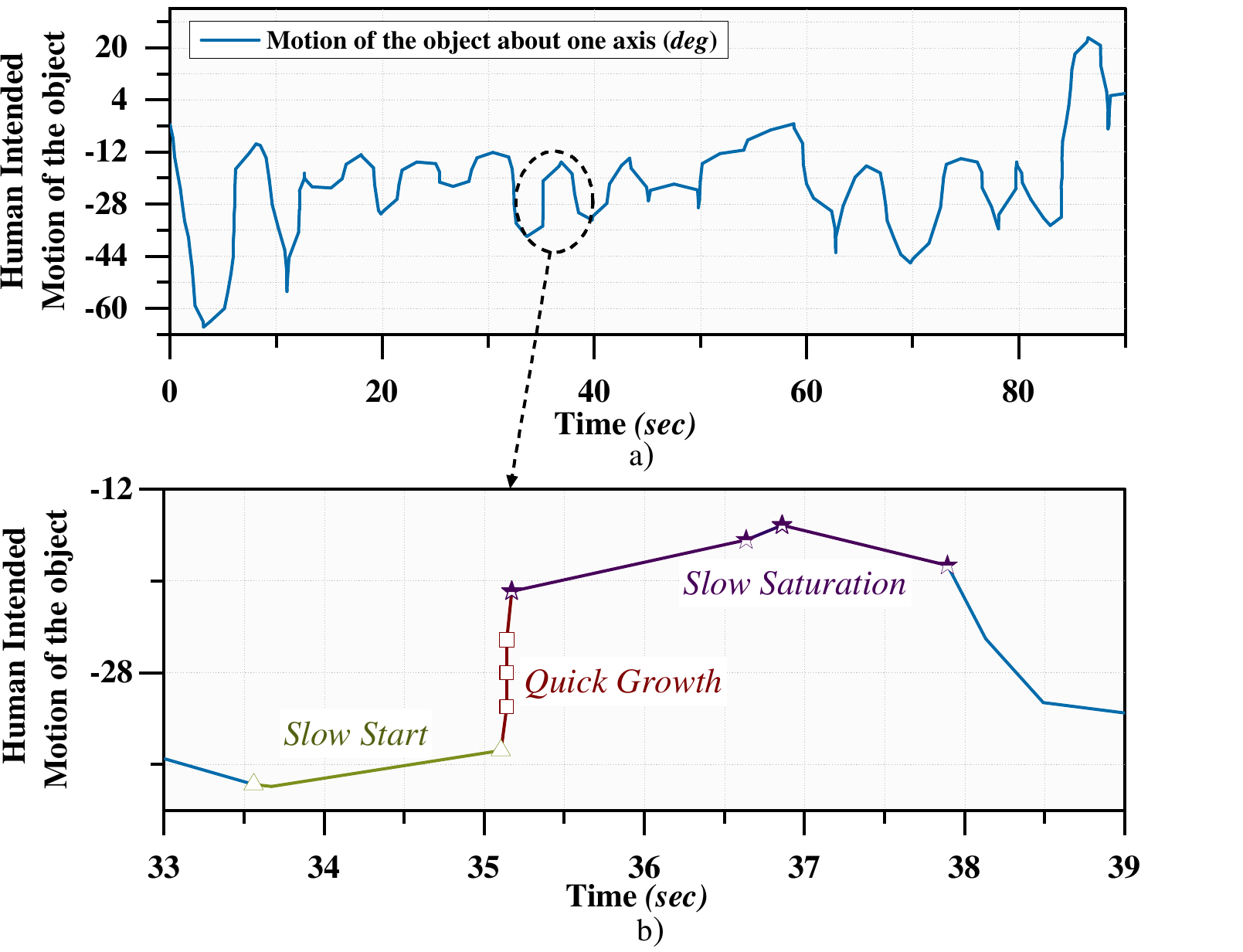}
    \vspace{-0.75cm}
    \caption{(a) Estimated human intended motion of the object (b) Detailed analyses of a single wavelet.}
    \label{intent_init}
    \vspace{-0.1cm}
\end{figure}

 $\mathcal{L}_\alpha(\hat{\theta}_\alpha,\theta_\alpha) = \frac{1}{d-m-r}\sum_{t=r}^{d-m-1} (\hat{\theta}_\alpha(t+m)-{\theta}_\alpha(t+m))^2$, where $d$, $m$, and $r$ denote the number of time distributed samples in the dataset, lookahead size and input window size, respectively. \begin{figure}[t]
    \centering
    \includegraphics[width =0.9\linewidth ]{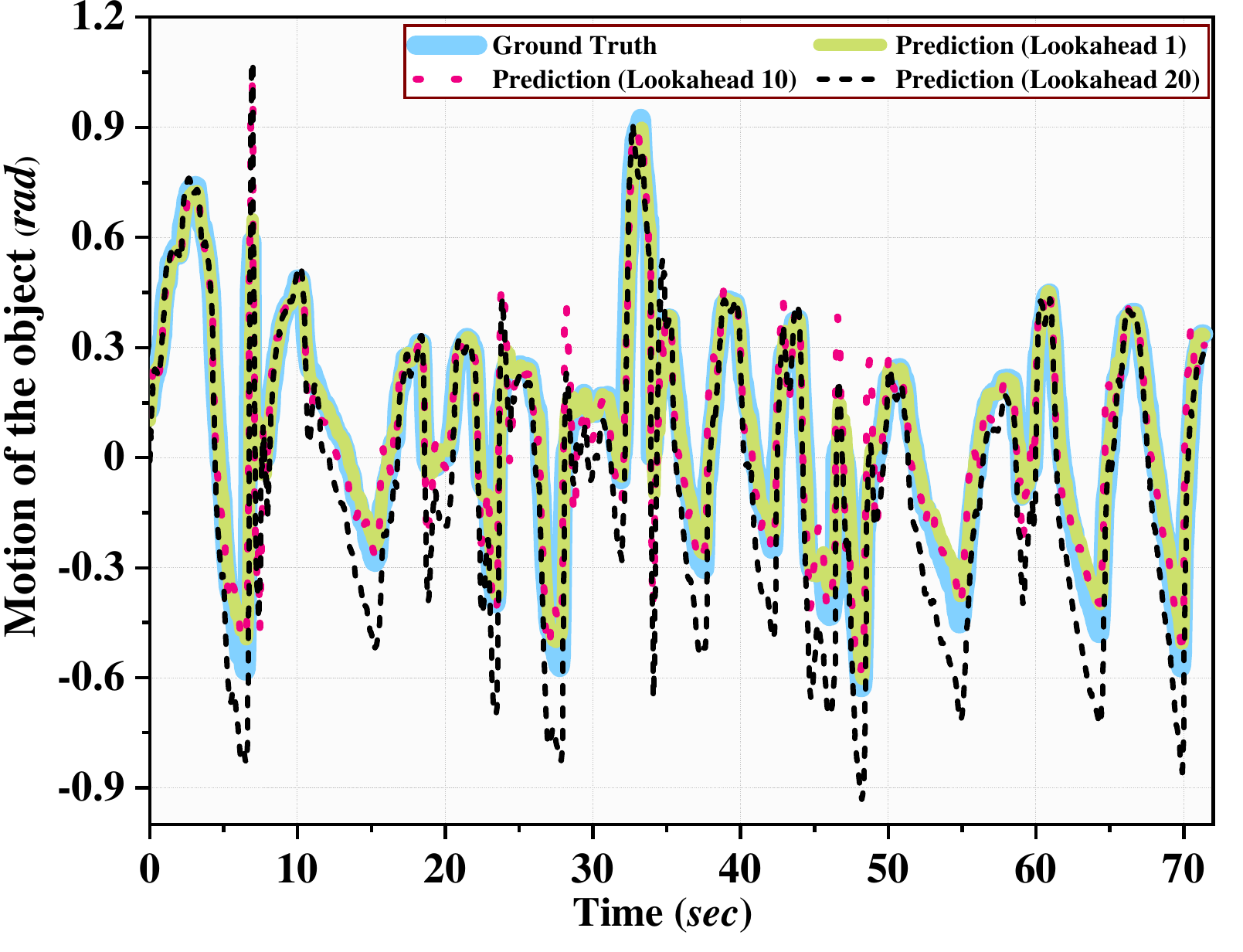}
    \caption{Predicted and actual trajectories of object motion using proposed attention model across different lookaheads.}
    \label{fig:TRANSFORMERRESULTS}
    \vspace{-0.4cm}
\end{figure} The test MSE about the three Cartesian axis for varying lookahead ($m$) = 1, 2, 3, 4, 5, 6, 7, 8, 9, 10, 15, 20, is reported as follows: a) $\mathcal{L}_x(\hat{\theta}_x,\theta_x)$ b)  $\mathcal{L}_y(\hat{\theta}_y,\theta_y)$ = c)  $\mathcal{L}_z(\hat{\theta}_z,\theta_z)$. The trajectories of the ground truth and the predicted estimates of intent across different lookaheads for a random manipulation of an in-held object is shown in Fig. \ref{fig:TRANSFORMERRESULTS}. It is observed that the mean squared error, across training, validation and test samples, increases with the increasing value of lookahead. Increasing error values are expected as a result of increase in the prediction window.
\\\indent\textbf{Observation of Human Behavior:} We present an analysis of the human behavior while manipulating objects with its fingertips. As shown in Fig. \ref{intent_init}, it can be seen that the motion intent of the human to manipulate the object follows a sigmoidal-like curve, consisting of a slow growth followed by a quick surge and then a slow saturation. The sequential variability in the estimated human intent is small at the initial stages, thereby leading to lag in the motion of the robot (combined with the slow initial response during this period) which is mainly observed because of following a move-and-wait strategy.
\newline\indent\textbf{Analysis of delays:} The relationship between the parameter $m$ and the observed error involves a trade-off. The length of the lookahead window ($m$) would account for an equal number of round-trip delays, but this results in an increase in error as the window size grows. The latency of the channel is influenced by network dynamics, which falls outside the scope of this study. However, we proceed to analyze the performance of predictive mechanism in mitigating the observed delays. The experimentation was conducted on real-world 5G-network in Robotic Operating System (ROS) environment, with an average round-trip latency of approximately $40\pm 5$ ms when the air-distance between haptic glove (at the master site) and the robotic hand (at the slave site) is $~9$ km. The cumulative delays in generating predictions/estimations for human intent ($1-2$ ms), control mechanism ($0.12-0.25$ ms), and processing/actuation delay ($10-30$ ms), excluding human reaction time, add up to $\sim 46.12-77.25$ ms, for a single round-trip. Hence, the end-to-end system utilizes $11.12-32.25$ ms to compensate for this delay that would otherwise occur in a singular round trip of control-feedback signals.
\\\indent\textbf{Discussion on generalization across users, objects and motion:} The proposed mechanism entails a closed-loop system, with the forward kinematics of the robot as feedback, it is reported to be robust to variations in the shape/pose of the objects. While as achieving generalization across human users is intrinsic to the design of haptic glove~\cite{gu2016dexmo}, we performed the manipulation of the three objects under study with 6 able-bodied human participants, aged between 18 and 55, out of which 3 participants had some prior experience with robot interaction. This experiment was approved by the Institute Ethics Committee of IIT Delhi.\par In this study, the aim is to quantify the intent of human motion in terms of the desired goal pose configuration of the object as it undergoes rotational motion within the grasp of a RH (controlled with a HG). Other macro-motions, including in-hand translation, could be represented as combinations of these rotational actions. Since, the estimated/predicted intent observes a continuous distribution $(\theta, \theta_{pr} \in \mathbb{R})$, it is generalized across the motion that the object can achieve while being held by the robotic hand. 
\section{Conclusion}
The study primarily focuses on establishing an estimation mechanism for transforming the joint motion signals captured from a human hand by an exoskeleton haptic glove to be characterized as the expected motion of the object being manipulated within the grip of the robotic hand. The challenges addressed are high dimensional mapping, lack of standardization, and inherent variability of kinematics in the devices. A control mechanism is introduced to transform the synthesized expected goal pose of the object into joint state configuration of the robotic hand.  In order to mitigate the delays occurring because of communication, control and processing, an attention-based CNN encoder is introduced to synthesize a prediction of the estimated intent of motion. The end-to-end system is evaluated on real-world robotic setup using Allegro Robotic Hand and Dexmo Haptic Glove in 5G communication environment, across different objects and human subjects, for analysing the delays, human behavior, performance, and generalizability. The proposed methodology has reported significant improvement in accuracy against existing studies and has the potential for deployment in a real-world scenarios.
\bibliographystyle{IEEEtran}
\bibliography{ref.bib}
\end{document}